# Device-friendly Guava fruit and leaf disease detection using deep learning

Rabindra Nath Nandi[1][0000-0002-9350-8043], Aminul Haque Palash[1][0000-0002-4794-8531], Nazmul Siddique[2][0000-0002-0642-2357], and Mohammed Golam Zilani[3][0000-0002-8783-327X]

[1] BJIT Limited, Dhaka, Bangladesh
{rabindro.rath,aminulpalash506}@gmail.com
[2] School of Computing, Engineering and Intelligent Systems, Ulster University, United Kingdom
nh.siddique@ulster.ac.uk
[3] Swinburne Online Australia, Australia
gzilani@swin.edu.au

**Abstract.** This work presents a deep learning-based plant disease diagnostic system using images of fruits and leaves. Five state-of-the-art convolutional neural networks (CNN) have been employed for implementing the system. Hitherto model accuracy has been the focus for such applications and model optimization has not been accounted for the model to be applicable to end-user devices. Two model quantization techniques such as float16 and dynamic range quantization have been applied to the five state-of-the-art CNN architectures. The study shows that the quantized GoogleNet model achieved the size of 0.143 MB with an accuracy of 97%, which is the best candidate model considering the size criterion. The EfficientNet model achieved the size of 4.2MB with an accuracy of 99%, which is the best model considering the performance criterion. The source codes are available at https://github.com/CompostieAI/Guava-disease-detection.

## 1 Introduction

Agriculture is a key player for sustainable development and global food security, which is becoming more challenging with the growing world population. Plant health monitoring is very important for the desired crop yield in agriculture. Plant disease is a major threat to food security and economic loss in agriculture every year. There are plant pathogens which can spread from plant to plant very fast. Therefore, early diagnosis of plant disease is of key importance to prevent disease spread and crop production. Pathological test for diagnosis of plant disease is very often time consuming due to sample collection, processing and analysis. Moreover, pathological laboratory facilities are also not available in many parts of the country. An alternative to pathological test is the traditional visual assessment of plant symptoms. The traditional method requires an experienced expert in the domain but the visual assessment method is too subjective and the domain expert may not always be available in remote areas. Moreover, there are variants of disease due to variations of plant species and climates, which can make it difficult for an expert to diagnose plant disease correctly. Computer vision and machine learning methods can be employed for plant disease diagnosis with higher accuracy [1-3].

Deep learning-based studies are increasing due to their promising result in numerous applications in agriculture and big data. A study shows that a Deep CNN (Convolutional Neural Network) provides an accuracy of 99.35% for a dataset containing 54,000 images of 14 crop diseases and 26 different types of diseases. Many of such web-based applications do not consider the end-users' device which leads to degrading the performance of the system [4]. A similar study with a dataset containing 7,000 images of 25 different plant categories is reported where multiple DCNN architectures are used and the highest accuracy achieved is 99.53%. The study suggested the use of the model for real-time plant disease identification [5]. Unfortunately, these works didn't mention the model size complexity and application overhead for real-time plant disease detection.

Deep architectures are usually large in size and the inference time is directly dependent on the number of parameters of the architecture. Sometimes a trained model with a reasonably large size is not feasible for mobile-based applications where hardware does not support the execution of the model in real-time [6]. Edge AI refers to the use of AI models for prediction in Edge devices and it is currently an active research area to use the highly efficient larger deep model on Edge devices [7]. The first requirement of Edge AI is model size optimization. The optimized models are smaller in size and suitable to deploy on Edge Devices. Google has some fascinating tools and techniques for model quantization and compression and lite version conversion [8].

This research investigates different deep learning models for fruit and leaf disease detection based on available dataset. Two popular quantization techniques are employed: 1) Float16 quantization and 2) Dynamic range quantization. The model performances are verified after quantization and found promising. The empirical investigations show that the optimized models are feasible for use by affordable smartphones with low specifications available in Bangladesh.

The key contributions of this research are: i) the development of deep learning models for fruits and leaves disease diagnosis, ii) optimization of deep learning models, and 3) conversion of the models into optimized TFlite version applicable to smartphone applications.

The rest of the paper is organized as follows: Section 2 describes related works, Section 3 describes the dataset, Section 4 describes the method, Section 5 presents the experiments and Section 6 concludes the paper with few directions for future work.

## 2 Related works

There are several studies done using both machine learning and deep learning for Guava disease detection. The generic procedure for detection of plant disease consists of five stages: image acquisition, image labeling, feature extraction and fusion, feature selection, and classification of disease. Image acquisition is the first step towards image processing. It is done with a high-resolution digital camera followed by labeling for classification. Feature extraction is a very crucial part that refers to the process of transforming raw data (i.e., image) into numerical features that can be used by a machine learning algorithm. The most widely used features are color features and Local Binary Pattern (LBP) [9], Gray Level Co-Occurrence Matrix (GLCM) [10], and Scale Invariant Feature Transform (SIFT) [11]. Image segmentation [12], a process of partitioning the image into multiple segments or set of pixels, is often applied before feature extraction. Deep convolutional neural network (DCNN) is used for disease detection. Bhushanamu et al. [13] used Temporal CNN where, firstly, contour detection is applied to detect the shape of the leaf and Fourier feature descriptor is used as features to the 1D CNN. Mostafa et al. [14] used five different CNN structures such as AlexNet, SqueezeNet, GoogLeNet, ResNet-50, and ResNet-101 to identify different guava diseases.

A CNN based plant disease identification model is developed by Mohanty et al. [4] which can detect 26 diseases of 14 crop species. An attention mechanism is developed by Yu et al. [15] that highlights the leaf area capable of capturing more discriminative features. Jalal et al. [16] used DNN to develop a plant disease detection system for apple leaf diseases where SURF is used for feature extraction and an evolutionary algorithm is used for feature optimization. A leaf disease model based on MobileNet model is developed in [17] and the performance of MobileNet is evaluated and compared with ResNet152 and InceptionV3 models.

In all of these studies, deep learning models are developed and trained for plant disease identification but there has been no proper study on the model optimization meaning reduction of model size suitable for end-user device applications.

## 3 Dataset

Guava belongs to the Myrtaceae plant family and it is a common tropical fruit cultivated in many tropical and subtropical regions Bangladesh, India, Pakistan, Brazil, and Cuba [14]. Guavas are incredibly delicious and rich in antioxidants, vitamin C, potassium, calcium, nicotinic acid, and fiber.

The dataset is based on Bangladesh and collected from a large Guava garden in the middle of 2021 by an expert team of Bangladesh Agricultural University. No pretreatment was applied before collecting the image using a Digital SLR Camera [18]. The dataset consists of four typical diseases: Phytophthora, Red Rus, Scab, Styler end Rot as well as Disease-free leaf and fruits. Phytophthora is a fruit disease caused by fungus and manifests as black lesions on young fruits. Red Rus is a Guava leaf disease caused by fungus. Scab is caused by fungi and has the symptoms of ovoid, corky, and round lesions on the surface of guava fruits. Styler end Rot begins at the styler and spreads to the root in guava fruits shown in Fig. 1.

The dataset has two types: original dataset and augmented dataset. Original dataset has 681 samples and the augmented dataset has 8525. Class-wise data distribution is provided in Table 1. The minimum samples are 87 of Red Rus disease and maximum number of samples are 154 belonging to Disease-free (fruit). It can be roughly said that the original dataset is a roughly balanced dataset. The augmented dataset is quite larger than the original dataset and about 10 times larger than the original dataset. The range of the augmented dataset is 1264 to 1626. For the sake of experiments, both datasets original and augmented are combined.

**Table 1.** Disease-wise data distribution for both original and augmented dataset

| Disease Name | Original Data | Augmented Data |
| --- | --- | --- |
| Phytophthora | 114 | 1342 |
| Red Rus | 87 | 1554 |
| Scab | 106 | 1264 |
| Styler end Rot | 96 | 1463 |
| Disease-free (leave) | 126 | 1276 |
| Disease-free (fruit) | 154 | 1626 |

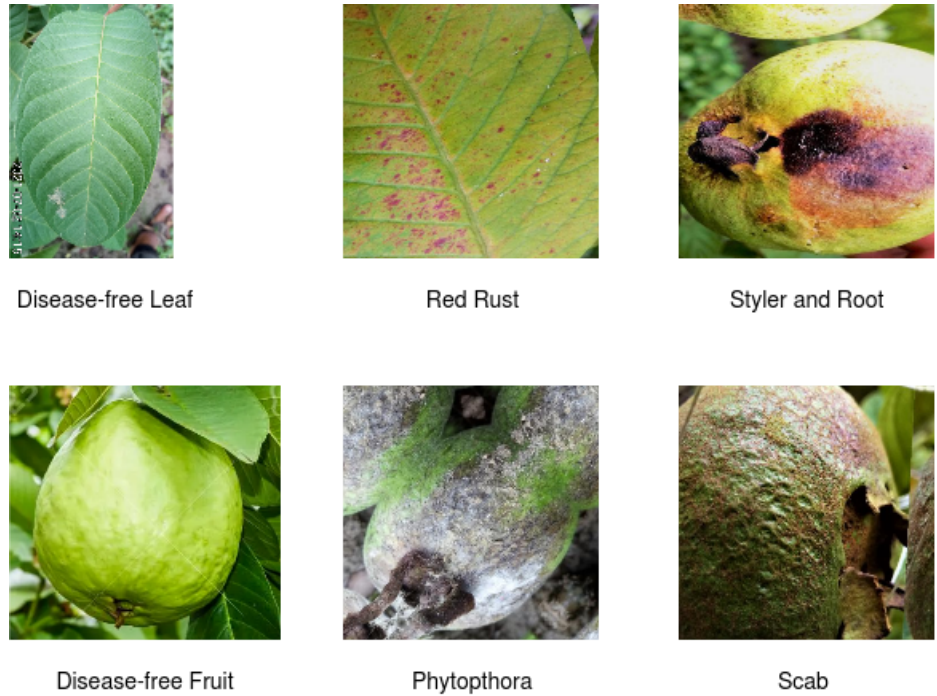

**Fig. 1.** Samples images with four types of diseases (Red Rust, Styler and Root, Phytopthora, Scan), Disease-free Leaf, and Disease-free Fruit.

## 4   Method

The main objective of this research is to understand the scope of model optimization with high detection performance for disease detection. The work comprises two parts: 1) model training and 2) model optimization. The overall architecture is illustrated in Fig. 2.

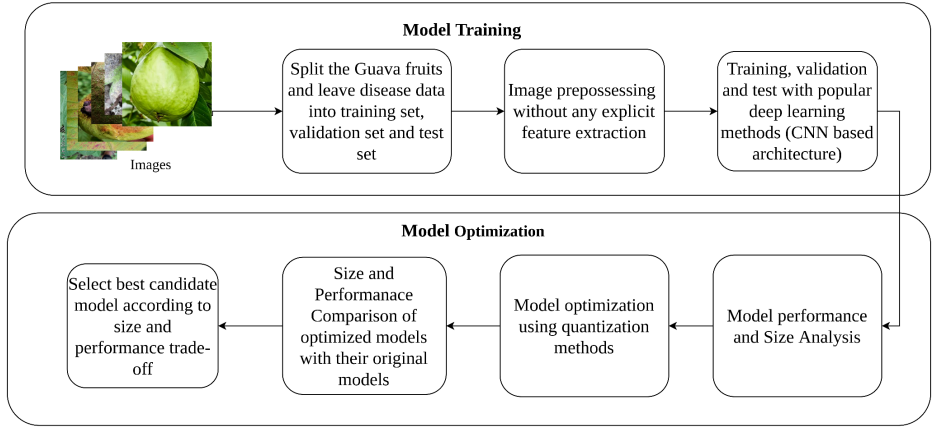

**Fig. 2.** The overall workflow of our system. Data split to model training and performance analysis, model optimization, and further performance analysis to the optimized model.

### 4.1 Model Training

Dataset, i.e., images, is firstly split into the training, validation, and test sets as source images are not provided in a grouped manner. Then, image resizing is applied as a part of image processing, and no special type of feature extraction is used before using the model.

Five prominent image classification models are used. These are VGG-16 [19], GoogleNet [20], Resnet-18 [21], MobileNet-v2 [22], and EfficientNet [23]. The model parameters are described in Table 1. VGG-16 has the highest number of parameters 138,357,544. MobileNet-v2 has 2,230,277 parameters which is the lowest in number among these models. A common experimental setup is used and pretrained models are used for fine-tuning to the Guava dataset. During the model training, the models are tested on both training and validation data, and after the model training, models are saved and validated with the test dataset.

**Table 2.** No of trainable parameters of different models on ImageNet dataset

| Model | Trainable Parameters |
|---|---|
| VGG-16 | 138,357,544 |
| GoogleNet | 5,605,029 |
| ResNet-18 | 11,179,077 |
| Mobilenet-V2 | 2,230,277 |
| EfficientNet-b2 | 4,013,953 |

The parameters of the listed model are large in number and model size grows according to the increase of the parameters which have an inverse impact on inference time, battery consumption and device storage. Hence, model optimization is needed to optimize and compress the model for suitable use in edge devices.

## 4.2 Model Optimization

Model optimization involves different factors: low latency, memory utilization, low power consumption, low cost, reducing payload size for over-the-air model updates. Especially on edge devices such as mobile and Internet of Things (IoT), resources are further constrained, and model size and efficiency of computation become a major concern. Beside this, inference efficiency is a critical concern when deploying machine learning models. One possible solution is to enable execution on hardware optimized-for fixed point operations and preparing optimized models for special purpose hardware accelerators.

According to Tensorflow documentation [24], there are three ways of model optimization such as: Quantization, Pruning and Clustering. Quantization reduces the precision of the numbers used to represent a model's parameters. Model parameters by default are 32-bit floating-point numbers. Pruning reduces the model parameters by removing parameters having a minor impact on model prediction. Clustering groups the weights of each layer in a model into a predefined number of clusters and then considering only the centroid values for the weights belonging to each individual cluster.

Only quantization techniques are applied to model optimization in this. There are mainly four types of quantization techniques: Post-training float16 quantization, Post-training dynamic range quantization, Post-training integer quantization, and Quantization-aware training [16]. Features of Post-training quantization techniques are described in Table 3.

**Table 3.** Details of three types of post-training quantization techniques

| Technique | Benefits | Hardware |
|---|---|---|
| Dynamic Range Quantization | 4x-smaller, 2x-3x speedup | CPU |
| Full Integer Quantization | 4x-smaller, 3x+ speedup | CPU, Edge TPU, Microcontrollers |
| Float16 Quantization | 2x smaller, GPU acceleration | CPU, GPU |

Post-training quantization details are depicted in Fig. 3. Float16 quantization reduces model sizes up to 50% by quantizing model constants (like weights and bias values) from full precision floating point (32-bit) to a reduced precision floating-point data type (IEEE FP16). Dynamic-range quantization reduces model size up to 75% by applying float-32 to float-8 quantization. Besides, it uses "dynamic-range" operators that dynamically quantize activations based on their range to 8-bits and perform computations with 8-bit weights and activations. Full Integer quantization requires calibration of all floating-point tensor ranges to get the max-min values. It requires a representative dataset to get the max-min values of variable tensors like model input, activations and model output. [16].

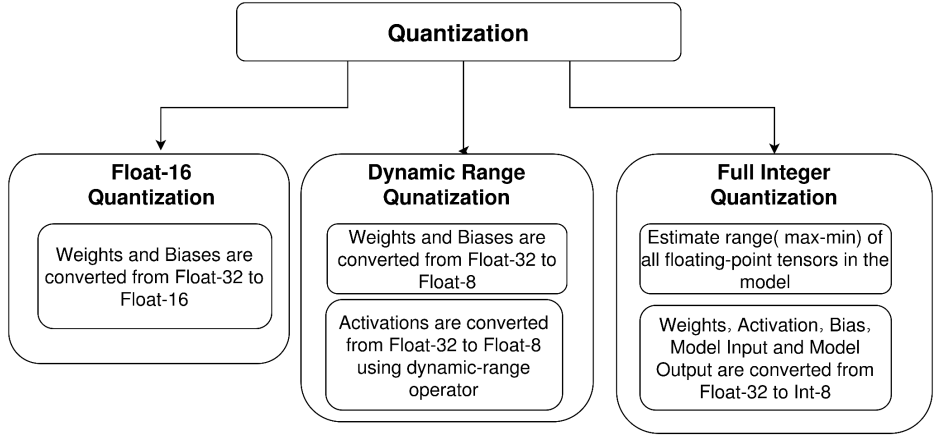

**Fig. 3.** 3-types of quantization techniques (Float-16 Quantization, Dynamic Range Quantization and Full Integer Quantization) with their operation details.

The quantized models are small in size compared to the original model. Theoretically, their performance should degrade also. To understand the effect of quantization, quantized models are tested with validation data and test data sets. The next step is to compare the quantized and original models according to both size and accuracy. Finally, to balance the trade-off between size and performance, an optimal model is chosen for specific hardware. The selected model is in TFlite format that can be directly used by Android and iOS applications with their convenient APIs.

## 5    Results and Discussion

Experiments have been carried out on all five models with pre-trained models without freezing the layers. Only the decision layer is changed as per the number of classes of guava disease. Stochastic Gradient Descent (SGD) is used with decay LR by a factor of 0.1 every 7 epochs. For evaluation purposes, accuracy, precision, recall, and macro-F1 scores are used.

   Table 4 shows the accuracy, precision, recall, and macro-f1 score for both validation data and training data. EfficientNet provides 99% accuracy and 100% f1-score which is the overall best result but other models also provide satisfactory results. GoogleNet provides 97% accuracy and f1-score on test data.

**Table 4.** Experimental results (Acc, Pr, Re, marco-F1) on training and testing data using different backbone cnn network

|  | Validation data | | | | Test data | | | |
| --- | --- | --- | --- | --- | --- | --- | --- | --- |
| Model | Acc | Pr | Re | F1 | Acc | Pr | Re | F1 |
| VGG-16 | 0.97 | 0.97 | 0.96 | 0.96 | 0.97 | 0.98 | 0.97 | 0.98 |
| GoogleNet | 0.96 | 0.96 | 0.96 | 0.96 | 0.97 | 0.97 | 0.97 | 0.97 |
| ResNet-16 | 0.96 | 0.97 | 0.96 | 0.96 | 0.97 | 0.97 | 0.97 | 0.97 |
| MobileNet-v2 | 0.97 | 0.98 | 0.98 | 0.98 | 0.98 | 0.99 | 0.99 | 0.99 |
| EfficientNet-b2 | 0.98 | 0.99 | 0.98 | 0.98 | 0.99 | 1 | 1 | 1 |

Table 5 shows the size reduction result after applying two types of quantization method. The VGG-16 model has the original size of 512.3 MB and is reduced to 256.1 MB and 128.01 MB after applying float16 quantization and dynamic range quantization respectively. This shows a reduction of model by 50% for float16 quantization and 75% for dynamic range quantization. For GoogleNet architecture, the original size of the model is 22.6 MB and the reduced size is 0.668 MB and 0.143 MB using float16 and dynamic range quantization respectively. This is about 156% reduction using dynamic range quantization. The current quantized model size is less than 1% of the original size. For ResNet-16 architecture, the reduction is from 44.8 MB to 22.4 MB and 1.7 MB for float16 quantization and dynamic range quantization respectively. This is about 96% reduction using dynamic range quantization. The current quantized model size is only 4% of the original size. For MobileNet-v2 architecture, the original model size is 9.2 MB and the quantized model size is 0.991 MB and 0.188 MB for float16 quantization and dynamic range quantization respectively. It is found that the MobileNet base model is small in size among all models compared to their original size. The reduction is 49 times smaller than the original model using dynamic range quantization. The quantized EfficientNet-b2 models have 8.1 MB and 4.5 MB where the original model size is 16.4 MB. It is shown that the dynamic range quantization method effectively reduces the model size for all five models compared to the Float-16 quantization. From Table 5, the quantized GoogleNet model size using dynamic range quantization is 0.143 MB, which is the lowest among all the models.

**Table 5.** Experimental size comparison before and after applying quantization

| Model | Optimization Method | Previous size (MB) | Optimized size (MB) |
| --- | --- | --- | --- |
| VGG-16 | Float16 quantization | 512.3 | 256.10 |
|  | Dynamic range quantization | 512.3 | 128.08 |
| GoogeNet | Float16 quantization | 22.60 | 0.668 |
|  | Dynamic range quantization | 22.60 | **0.143** |
| ResNet-16 | Float16 quantization | 44.80 | 22.40 |
| | Dynamic range quantization | 44.8 | 1.70 |
| MobileNet-v2 | Float16 quantization | **9.20** | 0.991 |
| | Dynamic range quantization | 9.20 | 0.188 |
| EfficientNet-b2 | Float16 quantization | 16.40 | 8.10 |
| | Dynamic range quantization | 16.40 | 4.50 |

From Table 6, it can be seen that the result doesn't change so much from the original model to the quantized models. Only 1-2% changes for a few models. VGG-16 model accuracy decreased from 97% to 96% and the F1-score from 98% to 96% for float16 quantization. For GoogleNet, there is no impact on model quantization and the accuracy is 97% for both original and quantized models. The accuracy of ResNet model is 97% and the quantized models have an accuracy of 96% and 95% respectively. For EfficeinetNet model, there is a minor change in accuracy, precision, recall and F1-score. From the performance perspective, EfficientNet model is the better among all quantized models. The reason for getting good performance of all models is probably the data quality is too good and only a few classes are available in the Guava dataset. Therefore, the optimal choice is to use the GoogleNet model using dynamic range quantization when the model size is the priority and if the model size is not an issue, then the EfficientNet model with dynamic quantization would be a good choice.

In this case study, model optimization techniques are explored and their impact on performance on storage size and memory are analyzed. It is clear that, if optimization hasn't been considered, MobileNet is the best choice as it has a size of 9.2 MB which is the lowest but after quantization. GoogleNet model with dynamic range quantization has size of only 0.143 MB and F1-score 0.97 is overall best candidate model from the all models described in Table 6. EfficientNet-b2 with dynamic quantization can be considered for mobile application if its size 4.5 MB is not a big issue as it has F1-score 0.99 which is better than GoogleNet quantized model.

**Table 6.** Experimental results (Acc, Pr, Re, marco-F1) on testing data using different backbone CNN network after applying model optimization and size comparison with original models

| Model | | | Test data | | | |
|---|---|---|---|---|---|---|
| | | Size | Acc | Pr | Re | F1 |
| VGG-16 | No optimization | 512.3 | 0.97 | 0.98 | 0.97 | 0.98 |
| | Float16 quantization | 256.10 | 0.96 | 0.96 | 0.96 | 0.96 |
| | Dynamic range quantization | 128.08 | 0.95 | 0.96 | 0.96 | 0.95 |
| GoogleNet | No optimization | 22.6 | 0.97 | 0.97 | 0.97 | 0.97 |

| Model | Optimization | Size (MB) | | | | |
|---|---|---|---|---|---|---|
| | Float16 quantization | 0.668 | 0.97 | 0.97 | 0.97 | 0.97 |
| | Dynamic range quantization | **0.143** | 0.97 | 0.97 | 0.97 | 0.97 |
| ResNet-16 | No optimization | 44.8 | 0.97 | 0.97 | 0.97 | 0.97 |
| | Float16 quantization | 22.4 | 0.96 | 0.96 | 0.96 | 0.96 |
| | Dynamic range quantization | 1.70 | 0.95 | 0.96 | 0.95 | 0.95 |
| MobileNet-v2 | No optimization | 9.20 | 0.98 | 0.99 | 0.99 | 0.99 |
| | Float16 quantization | 0.991 | 0.96 | 0.97 | 0.97 | 0.96 |
| | Dynamic range quantization | 0.188 | 0.96 | 0.97 | 0.97 | 0.96 |
| EfficeintNet-b2 | No optimization | 16.4 | **0.99** | 1.00 | 1.00 | 1.00 |
| | Float16 quantization | 8.10 | 0.99 | 0.99 | 0.99 | 0.99 |
| | Dynamic range quantization | 4.50 | 0.99 | 0.99 | 0.99 | **0.99** |

## 6  Conclusion and Future Work

Device end machine prediction is an active research area to overcome the complexity and cost of cloud computing. In this study, it is demonstrated that model optimization is an elegant way to use large-scale neural network models in edge computing. A similar study can be applied to other problems and datasets also. The future plan is to work with large and complex disease datasets not only considering the aspects of theoretical justification but also the real-life product development.